\documentclass{article}

\usepackage{arxiv}
\usepackage[utf8]{inputenc} 
\usepackage[T1]{fontenc}    
\usepackage{hyperref}       
\usepackage{url}            
\usepackage{booktabs}       
\usepackage{amsfonts}       
\usepackage{nicefrac}
\usepackage{microtype}
\usepackage{cleveref}
\usepackage{lipsum}
\usepackage{graphicx}
\usepackage{doi}
\usepackage{caption}
\usepackage{subcaption}
\usepackage{mwe}
\usepackage{hyperref}

\usepackage{times}
\usepackage{latexsym}

\newcommand{\carbondioxide}{CO\textsubscript{2}}
\newcommand{\carboneq}{CO\textsubscript{2}eq}
\newcommand{\carbonintensity}{gCO\textsubscript{2}eq/kWh}

\usepackage[utf8]{inputenc}
\usepackage{tablefootnote}

\usepackage{microtype}

\title{Estimating the Carbon Footprint of BLOOM, \\  a 176B Parameter Language Model}

\author{Alexandra Sasha Luccioni \\
        Hugging Face \\
  \texttt{sasha.luccioni@hf.co} \\\And
        Sylvain Viguier \\ 
        Graphcore \\
    \texttt{sylvainv@graphcore.ai} \\ \And 
    Anne-Laure Ligozat \\
 LISN \& ENSIIE \\ \texttt{anne-laure.ligozat} \\ \texttt{@lisn.upsaclay.fr}}

\renewcommand{\shorttitle}{Estimating the Carbon Footprint of BLOOM}

\date{}

\begin{document}

\maketitle

\begin{abstract}
Progress in machine learning (ML) comes with a cost to the environment, given that training ML models requires significant computational resources, energy and materials. In the present article, we aim to quantify the carbon footprint of BLOOM, a 176-billion parameter language model, across its life cycle. We estimate that BLOOM's final training emitted approximately 24.7 tonnes of~\carboneq~if we consider only the dynamic power consumption, and 50.5 tonnes if we account for all processes ranging from equipment manufacturing to energy-based operational consumption. We also study the energy requirements and carbon emissions of its deployment for inference via an API endpoint receiving user queries in real-time. We conclude with a discussion regarding the difficulty of precisely estimating the carbon footprint of ML models and future research directions that can contribute towards improving carbon emissions reporting.
\end{abstract}

\section{Introduction} \label{sec:intro}

Climate change is one of our generation's biggest challenges, impacting ecosystems and livelihoods across the world; estimating and reducing our carbon emissions is an important part of mitigating its impacts~\cite{masson2018global}. According to recent estimates, the global \carbondioxide~emissions of the information and communications technology (ICT) sector account for around 2\% of global \carbondioxide~emissions, but this figure is hard to estimate precisely given the distributed nature of global computing infrastructure~\cite{itu2020greenhouse,malmodin2018energy, ict2020}. The infrastructure used for training and deploying machine learning (ML) models contributes to this number, but the exact extent of this contribution is also unclear. In order to get a better grasp of the carbon footprint of the field, it is important to start systematically tracking the carbon footprint of ML models and algorithms and the main sources of emissions.

Large language models (LLMs) are among the biggest ML models, spanning up to hundreds of billions of parameters, requiring millions of GPU hours to train, and emitting  carbon in the process. As these models grow in size -- which has been the trend in recent years -- it is crucial to understand to also track the scope and evolution of their carbon footprint. 
The current study describes the first attempt to estimate the broader carbon footprint of an LLM, including the emissions produced by manufacturing the computing equipment used for its training, as well as the by model deployment via an API. The goal of our study is not to hone in on an exact number for the emissions produced, but to provide estimates of the relative contribution of step of the deployment process towards the overall emissions. 
We conclude with a discussion about the carbon emissions of different LLMs as well as the BigScience workshop overall, and propose directions for future work to both quantify and report these emissions. 

\section{Related Work} \label{sec:related}

There are different aspects of the environmental impact of computing in general and machine learning in particular that are relevant to our study; we briefly describe existing relevant work in the paragraphs below.

\paragraph{Empirical Studies on ML \carbondioxide~Emissions} Most of the existing work in this area has been done on estimating the \carbondioxide{} emissions incurred during model training. Starting with the seminal work of Strubell et al., who looked at the carbon footprint of training a Transformer model~\cite{strubell2019energy}, more recent studies have also looked at other model architectures and their ensuing emissions~\cite{patterson2021carbon, naidu2021towards}. Other studies have pursued a broader analysis of trends in terms of the energy requirements and \carbondioxide{} emissions of ML models in general~\cite{thompson2020computational, wu2021sustainable,patterson2022}. While some studies predict a growth in terms of carbon emissions of ML models~\cite{thompson2020computational}, others have predicted that emissions will shrink in coming years~\cite{patterson2022}; further work is therefore needed to get additional estimates from a broader variety of models and use cases. 

\paragraph{Tools for Estimating Carbon Impact} Another relevant research direction has pursued the development of tools for estimating the \carbondioxide{} emissions of training ML models, resulting in several tools created for this purpose. Some of these run in parallel to model training code and track its energy consumption and \carbondioxide{} emissions (e.g.~\cite{schmidt2021codecarbon,anthony2020carbontracker}), while others can be used post-training in order to produce a more high-level estimate of emissions (e.g.~\cite{lacoste2019quantifying}). However, these tools remain seldom used for reporting the \carbondioxide~emissions in ML publications, and a recent study has found that they vary significantly in terms of the estimates that they produce~\cite{bannour2021evaluating}.


\paragraph{Additional Factors} Complementary work has also been done on other contributions to the overall carbon footprint of ML, ranging from the carbon footprint of in-person versus virtual conference attendance~\cite{skiles2021conference} to the manufacturing of computing hardware~\cite{gupta2021chasing} as well as the life cycle analysis of the entire ML development and deployment cycle~\cite{ligozat2021unraveling} and the certification of ML systems according to their social and environmental impacts~\cite{gupta2020secure}. Increasingly, scholars have adopted a broader perspective on considering the environmental impacts of ML models, going above and beyond only the \carbondioxide{} emissions of model training and considering aspects such as equipment manufacturing and deployment~\cite{wu2021sustainable,kaack2022aligning}. However, there is still a need for a common approach in terms of estimating and comparing the carbon emissions of ML models which spans these different parts of the model life cycle.

\section{Background and Methodology}\label{background-methdology}

\subsection{The BLOOM Model}\label{sec:model}
The BigScience Large Open-science Open-access Multilingual Language Model (BLOOM) is a 176 billion parameter language model. It was trained on 1.6 terabytes of data in 46 natural languages and 13 programming languages as part of the \href{https://huggingface.co/bigscience}{BigScience workshop}, a year-long initiative that lasted from May 2021 to May 2022 and brought together over a thousand researchers from around the world. The BigScience workshop was granted access to the computing resources of the  Institut du développement et des ressources en informatique scientifique (IDRIS) of the Centre national de la recherche scientifique (CNRS) in France, which meant that model training was carried out on the Jean Zay computer cluster of IDRIS. We present some key numbers about BLOOM model training in Table~\ref{table:stats} below, and refer readers to~\cite{BLOOMmodelcard, le2022language} for additional information about model architecture and training.

\begin{table}[ht]
\begin{center}
\begin{tabular}{l|c}
\textbf{Total training time} & 118 days, 5 hours, 41 min \\ \hline
\multicolumn{1}{l|}{\textbf{\begin{tabular}[c]{@{}l@{}}Total number of \\ GPU hours \end{tabular}}} & 1,082,990 hours \\ \hline
\textbf{Total energy used} & 433,196 kWh \\ \hline
\textbf{GPU models used} & Nvidia A100 80GB \\ \hline
\multicolumn{1}{l|}{\textbf{\begin{tabular}[c]{@{}l@{}}Carbon intensity \\ of the energy grid\end{tabular}}} & 57 \carbonintensity \\ \hline
\end{tabular}\vspace{0.2in}
\caption{Key statistics about BLOOM model training -- for more details about our methodology, see Section~\ref{subsec:dynamic}.}
\label{table:stats}
\end{center}
\end{table}
\vspace{-0.7cm}
While training the model was the culmination of the BigScience project, many other efforts were needed to achieve this goal. This includes initiatives such as: data sourcing, collection and processing, tokenization, architecture engineering and evaluation. Additionally, in the months preceding the final BLOOM training, several smaller-scale experiments were launched in order to evaluate different model sizes and architectures, which helped converge on the final BLOOM architecture. 
In the results that presented in Section~\ref{sec:emissions}, we report the carbon emissions produced by the final 176B parameter BLOOM model, whereas the emissions of intermediate model training and evaluation carried out within the scope of the BigScience project are presented in Section~\ref{subsec-bigsciemissions}.

\subsection{Methodology} \label{sec:methodology}

\begin{figure}[h!]
\begin{center}
  \includegraphics[width=\linewidth]{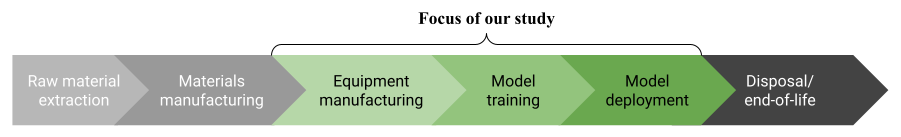}
  \caption{While the LCA approach encompasses all the stages of the product life cycle (from raw material extraction to disposal), we focus on those in green, which range from equipment manufacturing to model deployment.}
  \label{fig:cycle}
\end{center}
\end{figure}

While there is no universally-accepted approach for assessing the environmental impacts of ML models, we strive towards adopting the widely-used Life Cycle Assessment (LCA) methodology, which aims to cover all stages of the life cycle of a product or process~\cite{klopffer1997life}. While we do not have all of the necessary information to carry out a "cradle-to-grave" assessment of BLOOM (which would consider the environmental impacts of all processes from raw material extraction to disposal), we focus on the steps for which we do have sufficient information, which range from manufacturing the equipment used for training the model to model deployment (see Fig~\ref{fig:cycle}). In fact, recent work by Kaack et al. has proposed a more specific framework for categorizing ML's effects on greenhouse gas (GHG) emissions, consisting of 3 categories: (A) computing-related impacts, (B) immediate impacts of deploying ML and (C) system-level impacts on other domains~\cite{kaack2022aligning}. While this framework has not yet been widely adopted in our field, we believe it is particularly useful given the specificity of the ML life cycle. In the current study, we focus on category (A) and briefly discuss deployment and system-level impacts in Section~\ref{sec:futurework}.

Given that the LCA approach aims to account for all possible sources of GHG emissions (e.g. methane, carbon dioxide, nitrous oxide, etc.), it is necessary to convert these different gases to a single unit of measure in order to add them up. The standardized measure that is often used for this is \textit{carbon dioxide equivalents} (\carboneq{}), which are calculated based on comparing the global-warming potential (GWP) of different greenhouse gases to that of carbon dioxide (\carbondioxide). For instance, methane has a 100-year GWP 25 times that of \carbondioxide -- this means that it is equal to 25~\carboneq{}. In the sections below, we be using this unit of measure to estimate BLOOM's carbon emissions.

\section{Results: Carbon Emissions of the BLOOM Model} \label{sec:emissions}

Our study aims to bring together the different elements contributing to the overall carbon footprint of training BLOOM and to compare the relative contribution of each one towards BLOOM's total emissions. While we will predominantly focus on model training, we will also take into account the emissions produced by manufacturing the computing equipment used for running the training, the energy-based operational emissions, as well as the carbon footprint of model deployment and inference. All of the code and data used for our analyses are available in our \href{https://github.com/bigscience-workshop/carbon-footprint/}{Github repository}.

\subsection{Embodied Emissions} \label{subsec:manufacturing}
The \emph{embodied emissions} are those emissions associated with the materials and processes involved in producing a given product, such as the computing equipment needed to train and deploy ML models. While the production of these emissions is exclusively limited to the manufacturing process, this total amount is usually spread over the time during which equipment is used by dividing the total embodied emissions by the time of use. The BLOOM model was trained on HPE Apollo 6500 Gen10 Plus servers containing Nvidia A100 GPUs. The closest comparable computing equipment that provides LCA information is HPE's ProLiant DL345 Gen10 Plus server, which is similar to the Apollo 6500 and has a production footprint of approximately 2500 kg of ~\carboneq~\cite{hpe-proliant}. This does not include the embodied emissions of the GPUs which are used in the server, whose embodied emissions must be calculated separately. While Nvidia does not currently disclose the carbon footprint of its GPUs, recent estimates put the lower bound of this amount at approximately 150 kg of~\carboneq~\cite{davy-2021}, which is the number we will use for our embodied emissions estimates.

Assuming a replacement rate of 6 years and 85\% average usage (which are the figures provided to us by IDRIS), the figures above translate to an embodied carbon footprint of approximately 0.056 kg of~\carboneq~for each hour of server time and 0.003 kg of~\carboneq{} for each hour of GPU time. Given that BLOOM training lasted a total of 1.08 million hours using, on average, 384 GPUs across 48 computing nodes, we can estimate that the embodied emissions associated to BLOOM training represent approximately 7.57 tonnes for the servers and 3.64 tonnes for the GPUs, adding a total of \underline{11.2 tonnes of~\carboneq{}} to its carbon footprint. This does not include the embodied emissions of the rest of the computing infrastructure (e.g. the network switches, cooling equipment and other devices that power the network), which are difficult to quantify given that we do not have the necessary information regarding their distribution and usage. 

\subsection{Dynamic Power Consumption} \label{subsec:dynamic}

As described in Section~\ref{sec:related}, most of the existing research on the carbon footprint estimation of ML models has focused on estimating the~\carbondioxide~emissions produced by generating the electricity necessary for powering model training -- this is typically referred to as \textit{dynamic consumption}. This is calculated by multiplying the number of GPU hours used by the thermal design power (TDP) of those GPUs and the carbon intensity of the energy grid used to power the hardware. TDP remains an upper bound of GPU power consumption, but it is often used as a proxy given when access to real-time GPU power consumption is impossible. A grid's carbon intensity depends on the electricity source that powers it -- for instance, coal-powered grids result in more carbon emissions per kWh of electricity compared to grids powered by hydroelectricity or solar power. Also, while many compute providers carry out post hoc carbon offsetting or heat recycling, we do not take this into account in our estimation, given that we are focusing on the direct carbon emissions linked to dynamic power consumption~\footnote{In fact, a percentage of the heat produced by the Jean Zay computing cluster is recuperated to supply the heat and cold exchange network of the Paris-Saclay urban campus.}.

As reported in Table~\ref{table:stats}, training the BLOOM model required a total of 1.08 million GPU hours on a hardware partition constituted of Nvidia A100 SXM4 GPUs with 80GB of memory, which have a TDP of 400W~\cite{nvidia-a100}. While we were not able to track real-time power consumption, empirical observations noted that GPU utilization was typically very high, nearing 100\%. Also, we do not consider the power usage of CPUs, which consume approximately 40 times less energy than GPUs and which are typically not as solicited during the model training process. This represents an electrical consumption of 433,195 kWh of electricity during training; multiplied by the carbon intensity of the energy grid used, which is approximately 57~\carbonintensity~\cite{france-energy}, this results in a total of \underline{24.69 tonnes of \carboneq}~ emitted due to dynamic energy consumption. 

\subsection{Idle Power Consumption} \label{subsec:static}

So far, the emphasis in the ML community has been on estimating the energy consumption and ensuing carbon emissions of the energy used to power specialized hardware such as GPUs. However, it is important to keep in mind that the broader infrastructure that maintains and connects this hardware also requires large amounts of energy to power it -- this is referred to as \textit{idle consumption}. The quantity of energy needed for this depends on the efficiency of the computing cluster that is being used and the configuration of the devices on the cluster. This can be reflected in part by factoring in the PUE (Power Usage Effectiveness) of the data centers used for training these models, which is the approach adopted by Patterson et al. for estimating the carbon emissions of ML models such as T5 and GPT-3~\cite{patterson2021carbon}. However, while PUE is a useful metric for representing the amount of energy used for cooling and other overhead, it does not account for the totality of energy consumed by the data center infrastructure~\cite{brady2013case,kurpicz2018energy}. 

\begin{table}[ht]
\begin{center}
\hspace{-0.4cm}
\begin{tabular}{c|c|c}
\textbf{Computing Mode} & \multicolumn{1}{|l}{\textbf{\begin{tabular}[c]{@{}c@{}}Power consumption \end{tabular}}}  & \multicolumn{1}{|l}{\textbf{\begin{tabular}[c]{@{}c@{}} Percentage of total \end{tabular}}}  \\ \hline
Infrastructure consumption & 27 kWh & 13.5\% \\\hline   
Idle consumption & 64 kWh & 32\% \\\hline
Dynamic consumption & 109 kWh  & 54.5\% \\\hline
\textbf{Total consumption} & \textbf{200 kWh} & \textbf{100\%}\\
\end{tabular}\vspace{0.2in}
\caption{Breakdown of power consumption of the A100 partition of the Jean Zay cluster. \textit{Infrastructure mode} measures the power consumed by networking systems, datacenter maintenance and cooling systems (i.e., servers are turned off). \textit{Idle} measures power consumed by servers turned on but unused. \textit{Dynamic} measures power consumed by servers actively training BLOOM.}
\label{table:idle}
\end{center}
\end{table}

In order to estimate the idle consumption of the computing infrastructure that we used for training BLOOM, we ran a series of experiments to compare the total energy consumption of idle devices on the Jean Zay computing cluster (e.g. network, GPUs, storage, cooling/heating and computation nodes) to the total consumption of the same devices while running the model training code. As we show in Table~\ref{table:idle}, we found that for the A100 partition of the cluster used for training the model, in \emph{Infrastructure mode} (with the computing nodes turned off but the network, storage and cooling turned on), the power consumption was 27 kWh; in \emph{Idle mode} (with network, storage and compute nodes on, but no processes running), the power consumption was 64 kWh. During BLOOM training, power consumption averaged at over 109 kWh. This indicates that only around 54\% of the power consumption can be attributed to running the code (i.e. the dynamic power consumption described in Section~\ref{subsec:dynamic}), whereas the remaining 46\% is used for keeping the computing nodes on. Multiplying this by the total training time, this adds a further 256,646 kWh of idle power consumption on top of the dynamic power used for training BLOOM, and \underline{14.6 tonnes of~\carboneq{}} to the overall carbon footprint of model training.

\begin{table}[ht]
\begin{center}
\hspace{-0.4cm}
\begin{tabular}{l|c|c}
\textbf{Process} &  \multicolumn{1}{l}{\textbf{\begin{tabular}[c]{@{}c@{}} \carbondioxide emissions \\ (\carboneq) \end{tabular}}} &   \multicolumn{1}{|l}{\textbf{\begin{tabular}[c]{@{}c@{}}Percentage of \\ total emissions \end{tabular}}} \\ \hline
Embodied emissions & 11.2 tonnes & 22.2 \% \\\hline
Dynamic consumption & 24.69 tonnes & 48.9 \%\\\hline
Idle consumption & 14.6 tonnes & 28.9 \%\\ \hline
\textbf{Total} & \textbf{50.5 tonnes} & \textbf{100.00\%}
\end{tabular}\vspace{0.2in}
\caption{Breakdown of \carbondioxide~emissions from different sources of the BLOOM model life cycle}
\label{table:emissions}
\end{center}
\end{table}

 While it may seem excessive to add such a large overhead to BLOOM's carbon footprint, taking embodied and idle emissions into account is a much better reflection of the true emissions of model training than solely considering the dynamic consumption of GPUs, as it also considers the network overhead and larger computing infrastructure without which training cannot take place. The figures from Table~\ref{table:emissions} are similar to those provided in product carbon footprint estimations for computing equipment (such as the one for the HPE servers used in Section~\ref{subsec:manufacturing}~\cite{hpe-proliant}), which estimate that the embodied emissions account for approximately 20-30\% of life cycle emissions, whereas use (i.e. the emissions of both dynamic and idle consumption) are 70-80\% of the total footprint. However, our estimations thus far have only been limited to BLOOM training -- in the following section, we aim to go further by doing an case study analysis of the energy consumption and ensuing carbon emissions of model deployment.

\subsection{Deployment and Inference} \label{subsec:inference}

In order to attempt to estimate the carbon emissions incurred by deploying BLOOM, we ran the \href{https://github.com/mlco2/codecarbon}{CodeCarbon} tool~\cite{lacoste2019quantifying} on a Google Cloud Platform (GCP) instance with 16 Nvidia A100 40GB GPUs, where BLOOM was deployed via \href{https://huggingface.co/bigscience/bloom}{an inference API}, and tracked the energy usage of the instance over a period of approximately 18 days. The model received an average of 558 requests per hour, which were handled in real time (i.e. without any batching), for 230,768 requests in total. While this is not necessarily representative of all deployment use cases, it is an example of real-time deployment of LLMs in applications such as chatbots, where they are expected to respond to a constant, varying flux of user queries. It also provides a useful data point for starting to measure the carbon emissions of ML model inference, which has not been the focus of much research to date.

\begin{figure}[h!]
\begin{center}
  \includegraphics[width=0.85\linewidth]{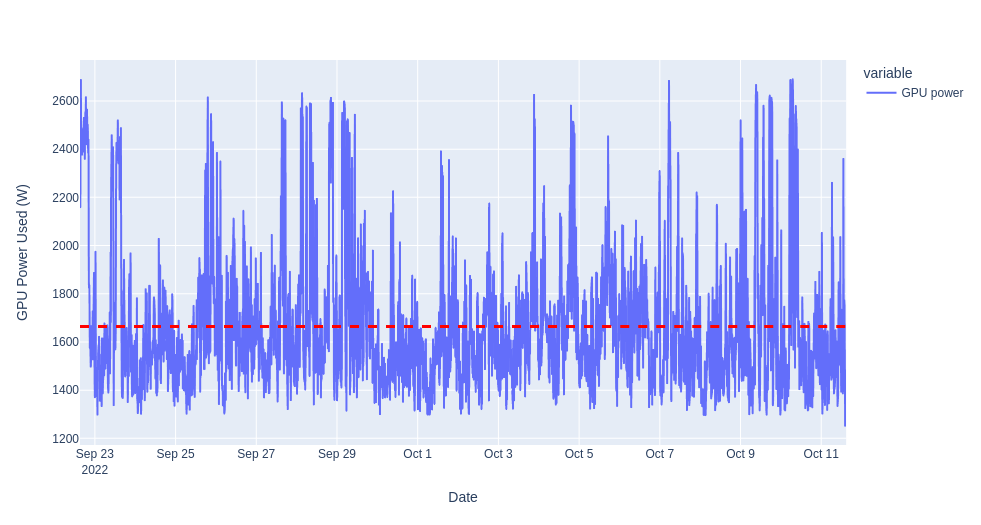}
  \caption{The fluctuation of mean power used to power the 16 Nvidia A100 GPUs running the BLOOM model API, with the mean power consumption in red (1664W) in dotted red.}
  \label{fig:power}
\end{center}
\vspace{-0.7cm}
\end{figure}

As it can be seen in Figure~\ref{fig:power}, during the 18 day period for which we carried out our analysis, the power consumed by the compute instance running the BLOOM model fluctuated between 1252 W to 2735 W -- divided by the 16 GPUs that were used, this amounts to 78-171W per GPU, which is significantly less than the TDP of this type of GPUs (400W). This indicates that the GPUs are not being used at maximum capacity, which is coherent with the nature of API deployment -- since inference requests are unpredictable, optimization of GPU memory using techniques such as batching and padding, which are the norm during training, is not possible, and the GPUs remain idle in between user requests.

\begin{figure}[ht!]
  \includegraphics[width=\linewidth]{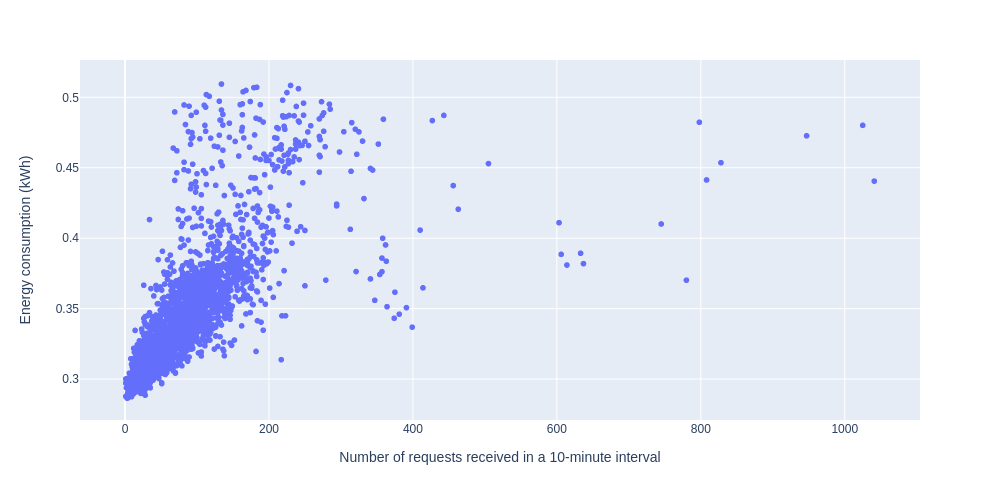}
  \caption{The quantity of energy used by the GCP instance (on the y axis) versus the number of requests received by the instance in a 10 minute interval (on the x axis). It can be seen that even when zero requests are received by the instance in this time span (bottom left of the graph), the energy consumption remains at approximately 0.28 kWh.}
  \label{fig:api}
\end{figure}

In total, the instance used for the BLOOM model API consumed 914 kWh of electricity -- of this amount, 22.7\% was consumed by the RAM (207.2 kWh), 2\% by the CPU (18.5 kWh) and 75.3\% by the GPU (688.38 kWh). It is hard to disaggregate this number into idle versus dynamic consumption because we do not have access to the GCP platform as we did for Jean Zay, but we can nonetheless  compare the energy consumed by the instance versus the number of requests that it received. We do so in Figure~\ref{fig:api}, where we plot the number of incoming requests to the BLOOM inference API and the energy consumption of the GCP instance where it is running. It can be seen in the Figure that even when there are almost no incoming requests during a 10 minute interval, there is still \(\sim\)0.28kWh of energy that is consumed during this interval, which represents the energy consumption of the model when it is not responding to any user requests. While more experimentation is needed in order to further disaggregate these numbers, we believe it is worth noting the high proportion of energy dedicated to maintaining a LLM like BLOOM in memory (approximately 75\% of the total energy consumed by the instance), without it being used. Going further, given that the GCP instance used for deploying the BLOOM model is running in the \texttt{us-central1} region, which has a carbon intensity of 394~\carbonintensity~\cite{gcp-carbon}, this results in approximately \underline{19 kgs of~\carboneq}~ emitted per day of API deployment, or 340 kg over the total period during which we were tracking emissions.

Given the many different combinations of configurations that can be used for deploying ML models, ranging from the hardware used for deployment to the batch size of inferences and the region where the model is running, the use case that we describe above is one among many. However,  it is a useful starting point to estimate the carbon emissions involved in deploying ML models, which are lacking in the field. While a 2019 article estimated that 80–90\% of Nvidia's ML workload is inference processing~\cite{nvidia-inference} -- a figure that was cited in a recent article by Patterson et al~\cite{patterson2021carbon}, a recent publication by Meta reported that inference accounted for approximately one-third of their end-to-end ML carbon footprint while the remainder owes to data management, storage, and training~\cite{wu2021sustainable}. We hope that the rough estimates we provide above shed some light on this question and plan on pursuing this avenue of research further in our future research endeavors, which we discuss in more detail in Section~\ref{sec:futurework}.

\section{Discussion and Future Work} \label{sec:discussion}

The main contribution of the present article is to define and connect the different sources of carbon emissions involved in training and deploying BLOOM, a 176B parameter language model. While we try to be as precise as possible in our calculations, they remain an estimate based on the information available and that which we have access to. In the current, final section of our article, we will compare our estimate to that of recent similar LLMs, attempt to estimate the dynamic consumption of all processes run within the scope of the BigScience workshop and discuss next steps and improvements that can be made to our approach to guide future work in the area. 

\subsection{Comparisons with other LLMs}

A few recent LLM papers reported the carbon footprint of model training, including notable models such as OPT-175B~\cite{zhang2022opt}, GPT-3~\cite{patterson2021carbon} and Gopher~\cite{rae2021scaling}. However, since the accounting methodologies for reporting carbon emissions are not standardized, it is hard to precisely compare the carbon footprint of BLOOM to that of these models. In this section, we will try to disentangle the different factors for each model: (1) the energy consumption of model training, (2) the \carboneq~emissions produced by dynamic consumption during training, and (3) the~\carboneq emissions produced via dynamic consumption while taking into account datacenter PUE (i.e. overhead) as well. We present these numbers in Table~\ref{table:comparison}, in which numbers in \textit{italics} indicate numbers that have been inferred based on the information provided in the papers accompanying these models, without being stated explicitly in articles and documentation.

\begin{table}[h]
\begin{center}
\begin{tabular}{c|c|c|c|c|c|c}
\hspace{-0.3cm}
{\textbf{\begin{tabular}[c]{@{}c@{}} Model \\ name \end{tabular}}}  &  \multicolumn{1}{l|}{\textbf{\begin{tabular}[c]{@{}c@{}}Number of \\ parameters \end{tabular}}}  & \multicolumn{1}{c|}{\textbf{\begin{tabular}[c]{@{}c@{}} Datacenter \\ PUE \end{tabular}}}&\multicolumn{1}{c|}{\textbf{\begin{tabular}[c]{@{}c@{}} Carbon intensity \\  of grid used \end{tabular}}} & \multicolumn{1}{c|}{\textbf{\begin{tabular}[c]{@{}c@{}} Power \\ consumption \end{tabular}}} & \multicolumn{1}{c|} {\textbf{\begin{tabular}[c]{@{}c@{}} \carboneq~\\ emissions  \end{tabular}}} & \multicolumn{1}{c}{\textbf{\begin{tabular}[c]{@{}c@{}} \carboneq~\\  emissions $\times$ PUE \end{tabular}}} \\ \hline
GPT-3 & 175B & 1.1 & 429~\carbonintensity & 1,287 MWh & \textit{502 tonnes}  & 552 tonnes\\ 
Gopher & 280B & 1.08 & 330~\carbonintensity & \textit{1,066 MWh} & \textit{352 tonnes} & 380 tonnes \\
OPT & 175B & \textit{1.09}~\footnote{Source: \cite{meta2021}} & \textit{231\carbonintensity} & \textit{324 MWh} &  70 tonnes &  \textit{76.3 tonnes~\footnote{By contacting the authors of the OPT paper, we were able to establish that they did not consider datacenter PUE in their carbon footprint estimation; however, we do not have the necessary information to accurately do so ourselves.}}  \\ 
BLOOM & 176B & 1.2 & 57~\carbonintensity & 433 MWh & 25 tonnes  & 30 tonnes \\ 
\end{tabular}\vspace{0.2in}
\caption{Comparison of carbon emissions between BLOOM and similar LLMs. Numbers in \textit{italics} have been inferred based on data provided in the papers describing the models.}
\label{table:comparison}
\end{center}
\end{table}

We can see that BLOOM training resulted in less than half of the emissions of the closest comparable model, OPT (which emitted 70 tonnes compared to BLOOM's 25 tonnes), and 20 times less than GPT-3 (502 tonnes). This can be explained in large part by the carbon intensity of the energy source used for training, given that the carbon intensity of the electric grid powering Jean Zay is 57~\carbonintensity, compared to 231~\carbonintensity{} for OPT, 429~\carbonintensity{} for GPT-3 and 330~\carbonintensity{} for Gopher. Comparing the raw energy consumption of the models is interesting as well because we can see that BLOOM actually consumed slightly more energy than OPT -- 433 MWh compared to OPT's 324 MWh, despite their proximity in size and training set up. Of course, there are also other factors that should be considered when comparing the energy consumption of models, such as the type of hardware used, the number of tokens seen by the model, the model architecture, etc., so an exact comparison is difficult, and it is useful to consider all of the characteristics described above when comparing models.

Finally, as we mentioned in Section~\ref{subsec:static}, the carbon footprint accounting approach proposed by Patterson et al.~\cite{patterson2021carbon} includes datacenter PUE, which is not always taken into account by other models. In order to allow a fair comparison, we attempt to disaggregate model carbon emissions with and without taking PUE into account in Table~\ref{table:comparison}. Since the PUEs of datacenters used for training ML models are relatively efficient and very similar (ranging from 1.08-1.2), their contribution to the overall carbon footprint of model training is relatively small.  However, as we have shown in Section~\ref{sec:emissions}, these numbers represent a small part of the actual carbon emissions and environmental impacts of training ML models, given that the reflect neither the embodied emissions nor the emissions due to model inference and deployment. In the next section, we attempt to go one step further by estimating the carbon footprint of intermediate experimentation and evaluation processes run within the scope of the BigScience workshop and how they compare to that of training the final BLOOM model.

\subsection{Carbon Footprint of the BigScience Workshop} \label{subsec-bigsciemissions}

The training of the 176B parameter BLOOM model represents only part of the experiments that were run on the Jean Jay computing cluster as part of the BigScience workshop. In fact, if we consider the totality of experiments run by members of the BigScience project, they add up to a total of 3.46 million GPU hours (2.2 million hours of which used V100 GPUs and 1.24 million hours used A100 GPUs), which represents an electrical consumption of 1,163,032~kWh of electricity and approximately 66.29 tonnes of \carboneq~emitted via dynamic power consumption. We break down this total into its different components, including the final BLOOM training, in Table~\ref{table:models}, below.

\begin{table}[ht]
\begin{center}
\hspace{-0.4cm}
\begin{tabular}{l|c|c|c}
\textbf{Process} &   \multicolumn{1}{l|}{\textbf{\begin{tabular}[c]{@{}c@{}}Energy consumed \\ (kWh) \end{tabular}}} &   \multicolumn{1}{l}{\textbf{\begin{tabular}[c]{@{}c@{}} \carbondioxide emissions \\ (tonnes of~\carboneq) \end{tabular}}} &   \multicolumn{1}{|l}{\textbf{\begin{tabular}[c]{@{}c@{}}Percentage of \\ total emissions \end{tabular}}} \\ \hline
176B BLOOM Model & 433,196  & 24.69 & 37.24\% \\\hline
104B Model  & 266,522 & 15.19 & 22.92\%\\\hline
1B Model & 158,972 & 9.06 & 13.68\%\\ \hline
13B Model & 87,210 & 4.97 & 7.49\%\\ \hline
Other Models & 64,257 & 3.66 & 5.53\% \\ \hline
Miscellaneous Processes & 57,961 & 3.3 & 4.98\% \\ \hline
6B Model & 51,686 & 2.95 & 4.45\% \\ \hline
Model Evaluation & 43,172 & 2.46 & 3.71\% \\ \hline
\textbf{Total} & \textbf{1,163,088} & \textbf{66.29} & \textbf{100.00\%}
\end{tabular}\vspace{0.2in}
\caption{Breakdown of dynamic energy consumption and \carbondioxide~emissions of different parts of the BigScience project}
\label{table:models}
\end{center}
\end{table}

It is interesting to note that experimenting with intermediate models (such as the 104B, 13B and 1B models) add up to a total of 35.8 tonnes of \carboneq, which is more than the training of the final model.  This is slightly higher than the estimate made by the authors of the OPT paper, who stated that the total carbon footprint of their model is roughly 2 times higher due to experimentation, baselines and ablations~\cite{zhang2022opt}. However, training these models allowed us to converge on the architecture and hyperparameters of the final BLOOM model, and many of these intermediate models were also shared with the community (e.g. \href{https://huggingface.co/bigscience/bloom-1b7}{BLOOM 1B} and \href{https://huggingface.co/bigscience/bloom-3b}{BLOOM 3B}). Other processes that contributed to the overall carbon emissions of the workshop included model evaluation, which emitted 2.46 tonnes of~\carboneq, as well as miscellaneous processes such as benchmarking, data processing and tokenization (3.3 tonnes of~\carboneq). While these processes are not part of the training of the model itself, we believe that it is important to estimate and report them as part of the research and development process -- we touch upon this point further in Section~\ref{sec:futurework}

If we take into account the embodied emissions of these processes, given that we used a total of 3.46 million GPU hours, this amounts to a total 35.9 tonnes of~\carboneq{}. Adding the embodied emissions and the idle consumption of equipment (according to the percentage described in Section~\ref{subsec:static}, this accounts for 73.32 further tonnes of~\carboneq, bringing up the total tally up to 123.82 tonnes of~\carboneq. Furthermore, given that these additional experiments also produced several LLMs that were shared with the community and deployed, they also continue to generate carbon emissions during their deployment and usage, which we are unable to account for but keep in mind as a further addition to our estimation.

\subsection{Future Work} \label{sec:futurework}

We hope that the present article shed some light on the different sources of carbon that contribute towards an ML model's total carbon footprint and how they compare. There are, however, many unanswered questions that we are lacking the data to pursue, some of which we will enumerate in the current section.

\paragraph{Gathering more precise figures regarding embodied emissions.} We used the closest available figures to compute the embodied emissions of manufacturing the GPUs used for training BLOOM. However, we were unable to get the figures for the exact hardware we are using, which makes the numbers we report an estimate. More transparency is needed regarding the environmental impacts of manufacturing computing equipment given the large quantities of chemicals and minerals required~\cite{stephens2013development,crawford2021atlas} and the significant quantities of ultra-pure water and energy needed to manufacture it~\cite{united2013geo}, as well as the complex and carbon-intensive supply chains and transportation involved in shipping them around the world~\cite{berkhout2004materialising}.

\paragraph{Running additional studies on model inference and deployment.} The results we report in Section~\ref{subsec:inference} barely scratch the surface of the complexity involved in deploying, scaling and maintaining ML models in practice and in real-time. We recognize this complexity but believe that bridging the gap between chip designers and users can help address this, as well as running more empirical studies to test different hardware setups and configurations and how they impact energy consumption and carbon emissions. 

\paragraph{Advocating for increased transparency and granularity in carbon reporting.} While some papers introducing ML models have begun reporting \carbondioxide~emissions, which we applaud, we also believe that disaggregating this single figure into aspects such as energy consumption, carbon intensity, and PUE is needed to allow for more meaningful comparisons between models. Furthermore, tallying and reporting the carbon emissions attributable to research and development, as well as evaluation and benchmarking, is useful to contextualize the relative contribution of the final model training towards that number.

\paragraph{Considering the broader impacts of ML.} In the existing research, the environmental impacts of information and communications technologies are classified into 3 categories: computing-related impacts due to the manufacturing of hardware and devices as well as electricity consumption; indirect impacts of deploying the models, and system-level impacts on other domains~\cite{kaack2022aligning}. In the current article, and all those we discussed in Section~\ref{sec:related}, the focus is put solely on the direct impacts of ML models. However, it can be useful to also consider their indirect impacts on industries such as transportation, agriculture or urban planning, given increased reliance on ML technologies in these sectors. We also do not discuss ML's impact on changing consumer behaviors, for instance more usage of devices such as smart speakers or connected devices to carry out tasks which were previously done by hand. While these impacts are harder to quantify precisely, we believe that they are nonetheless worth including in the broader environmental impacts of ML.

\section*{Acknowledgements}
We would like to thank the following people for their help and guidance in writing this paper:  Christopher Akiki, Clement Delangue,  Priya Donti, Udit Gupta, Lynn Kaack,  Remi Lacroix, Pierre-Francois Lavallée, Teven Le Scao, Nicolas Patry, David Rolnick, Thomas Wang, and the other members of the BigScience workshop. This work was granted access to the HPC resources of IDRIS under the allocation 2022-A0101012475 made by GENCI.

\bibliography{bibliography}
\bibliographystyle{acm}

\end{document}